\documentclass[10pt,twocolumn,letterpaper]{article}

\usepackage{cvpr}
\usepackage{times}
\usepackage{epsfig}
\usepackage{graphicx}
\usepackage{amsmath}
\usepackage{amssymb}

\usepackage[nolist]{acronym}
\usepackage{placeins}
\usepackage{wrapfig}
\usepackage{array}
\usepackage{amsthm}
\usepackage{algorithm}
\usepackage[noend]{algpseudocode}
\usepackage{makecell}
\usepackage{multirow}

\newcommand{\weight}{\mathbf{w}}
\newtheorem{assumption}{Assumption}

\PassOptionsToPackage{hyphens}{url}

\usepackage[pagebackref=true,breaklinks=true,letterpaper=true,colorlinks,bookmarks=false]{hyperref}

\cvprfinalcopy 

\ifcvprfinal\fi

\begin{acronym}
\acro{DNN}{Deep Neural Network}
\acro{SD}{Standard Deviation}
\acro{SGD}{Stochastic Gradient Descent}
\acro{CNN}{Convolutional Neural Network}
\acro{PRA}{Polyak-Ruppert Averaging}
\acro{BN}{Batch Normalization}
\acro{SWA}{Stochastic Weight Averaging}
\acro{PSWA}{Periodically Sampled Weight Averaging}
\acro{PWALKS}{Periodic Weight Averaging over Last K Samples}
\acro{PSWM}{Periodically Sampled Weight Momentum}
\acro{SWM}{Sampled Weight Momentum}
\end{acronym}

\newtheorem{mydef}{Definition}
\newtheorem{mytheorem}{Theorem}

\begin{document}

\title{Improving Model Training by Periodic Sampling over Weight Distributions}

\author{Samarth Tripathi\\
LG Advanced AI Lab\\
Sanat Clara, CA, USA\\
{\tt\small samarthtripathi@gmail.com}
\and
Jiayi Liu\\
LG Advanced AI Lab\\
Sanat Clara, CA, USA\\
{\tt\small jiayi.uiuc@gmail.com}
\and
Unmesh Kurup\\
LG Advanced AI Lab\\
Sanat Clara, CA, USA\\
{\tt\small unmesh.kurup@lge.com}
\and
Mohak Shah\\
LG Advanced AI Lab\\
Sanat Clara, CA, USA\\
{\tt\small mohak.shah@lge.com}
\and
Sauptik Dhar\\
LG Advanced AI Lab\\
Sanat Clara, CA, USA\\
{\tt\small sauptik.dhar@lge.com}
}

\maketitle

\begin{abstract}
In this paper, we explore techniques centered around periodic sampling of model weights that provide convergence improvements on gradient update methods (vanilla \acs{SGD}, Momentum, Adam) for a variety of vision problems (classification, detection, segmentation). Importantly, our algorithms provide better, faster and more robust convergence and training performance with only a slight increase in computation time. Our techniques are independent of the neural network model, gradient optimization methods or existing optimal training policies and converge in a less volatile fashion with performance improvements that are approximately monotonic. We conduct a variety of experiments to quantify these improvements and identify scenarios where these techniques could be more useful.
\end{abstract}

\section{Introduction}\label{sec:introduction}

Optimizing \acp{DNN} is especially challenging due to the nonconvex nature of their loss function. Hence, the development of gradient-based methods that use back-propagation to approximate optimal solutions has been crucial for neural network adoption. Optimization techniques over gradient updates like \ac{SGD} or gradient-based adaptive optimizers have made the training process more effective. However, optimal convergence of the loss function is still time-consuming, volatile, and needs many finely tuned hyperparameters. In this paper we show that by manipulating the model weights directly using their distributions over batchwise updates, we can achieve significant improvements in the training process, and add more robustness to optimization with negligible cost of additional training time.  Since our technique modifies the model weights directly using their distribution over gradient updates, it remains independent of gradient optimization methods and the model architecture.

Using the model weight distribution to achieve improvements on either the training process or a trained model has been widely studied by extending the \ac{PRA} method. ~\cite{2012arXiv1212.2002L} explored many techniques to speed up convergence of convex functions using the projected stochastic subgradient method. Their work explored gradient-based averaging, weighted averaging, and other variations, as well as the theoretical justifications for such an approach. ~\cite{NIPS2011_4316} explored how \ac{PRA} on \ac{SGD} has better convergence guarantees, especially when the initial condition on the weights are carefully removed from the averages and the learning rate is decayed correctly. However, most of this earlier research was focused on a theoretical understanding of weight-averaging methods and lacks analysis of their practicality especially on applications to highly nonlinear \ac{DNN} models.

Recently, similar techniques have been applied over model weight distributions, but mostly on pretrained models, (\cite{izmailov2018averaging} and \cite{liu2018res}). The techniques show better generalization and achieves a wider local minima post sampling. However, when such \ac{PRA} based methods are directly applied to train a \ac{DNN} from scratch, they fail to produce optimal performance. Meanwhile, these approaches also increase the computation load leading to increased training time. We investigate the reasons for the sub-optimal convergence for such methods, and address it by introducing PSWA. PSWA retains the robustness from regularization allowed by the \ac{PRA} techniques while mitigating their convergence problems. We further improve PSWA's shortcomings with PSWM and PWALKS, while also reducing the computational overhead substantially. These proposed techniques provide improvements to the training process of neural networks without compromising on optimal convergence.

The paper is organized as follows: we first present Related Works and highlight the problem common to previous research in the domain - the inability to converge optimally when training a \ac{DNN} model from scratch. In the following Methods section we propose three new techniques: \acs{PSWA}, \acs{PWALKS}, and \acs{PSWM} which address the major flaws with previous works. The Experiments section explores the application of our techniques with an empirical study using an adaptive optimizer in Adam and a more extensive Image dataset in ImageNet. In the Results section we analyze and quantify the improvement of our techniques followed by a Discussion section which explores the significance of our work and ends with the Conclusion.

\section{Related Work}\label{sec:prior}
\subsection{Methods}

Given a \ac{DNN} model with a loss function, $l(\weight, d_i)$, on a training sample $d_i$, the mini-batch \ac{SGD} method aims to minimize the expected loss $ \mathop{{}\mathbb{E}} _{d_i in D} [ l(\weight, d_i) ]$ of the training data ($D$) by updating the model weights $\weight$ iteratively as:

\begin{equation}\label{eq:update}
\weight^{(t+1)} = \weight^{(t)} - \eta\nabla_\weight \sum_{i \in \text{batch t} } l(\weight^{(t)}, d_i).
\end{equation}

The partial derivatives of the loss correspond to the direction of the gradient ascent of a batch of training data. The hyperparameter $\eta$ is the learning rate that controls the step size of the update. Most research in this area focus on the effects of different learning rate schedules, gradient update techniques algorithms, optimal batch sizes, etc. and how improvements in these areas can provide better convergence and add robustness \cite[as examples]{duchi2011adaptive,tieleman2012lecture,kingma2015adam,smith2017don}. These works are mainly dominated by the modified versions of the update $\eta \nabla_\weight l$. 

In comparison, the weight averaging approach aims to reassign the final value of weights as
\begin{equation}\label{eq:pra}
\weight_\text{final} = \frac1n\sum_{t=1}^{n} \weight^{(t)}
\end{equation}
from a sample of weights after $n$ batch updates. Variations on the application of this technique have been studied previously. ~\cite{izmailov2018averaging} proposed the \textit{\ac{SWA}} method, which uses \ac{PRA} over model distribution when retraining pretrained models to achieve flatter minimas and better generalization. This technique provides better generalization when \textit{finetuning} a model. ~\cite{2012arXiv1212.2002L} presented the \textit{Projected Stochastic Subgradient} method where an iteration-based weighed averaging approach to model training and its variations are explored. They presented theoretical analysis of the technique and discuss the finite variance bounds of their approach for SVM models.

\subsection{Challenges}

Rather than simply discuss their short-comings, we show empirically the failings of two salient previous works using the well-adopted ResNet18 ~\cite{he2016deep} on Cifar10 dataset ~\cite{krizhevsky2009learning}. We used the publicly available implementation \footnote{Refer \url{https://github.com/kuangliu/pytorch-cifar/blob/master/main.py}.}  using \ac{SGD} updates (with momentum and L2
weight penalty) and stepwise learning rate decay presented in Experiments section. Following the \ac{SWA} algorithm, we initialized a running mean for all model parameters and after training it for `c' epochs (`c'is a pre-defined hyperparameter), we replaced the model weights with their respective running means. Note that the running mean is initialized only once. It is then kept updated after each epoch and reassigned after `c' epochs, consistent with the original algorithm. We emphasize that the \textit{\ac{SWA}} technique we implement for baseline comparisons is a modification of the original implementation, to accommodate training models from scratch. We investigated two variations of ~\cite{2012arXiv1212.2002L} weighed averaging techniques on same experimental configurations. In the first approach `BachEpoch' we update the mean estimation after each epoch, with a weight of the $epoch$ value, and then reassign the mean values to those weights. In the second approach `BachBatch' we update the mean estimation after each batch, with a weight of cumulative batch total, and reassign model weights at the end of the epoch. Hence `BachEpoch' provides a linear weighed averaging approach and `BachBatch' provides an exponential weighed averaging approach. We also calibrate the \ac{BN} layers as mentioned in \cite{izmailov2018averaging}, by performing a forward pass over the training data after each reassignment, for all approaches.

\begin{figure}[th]
\centering
\includegraphics[width=\columnwidth]{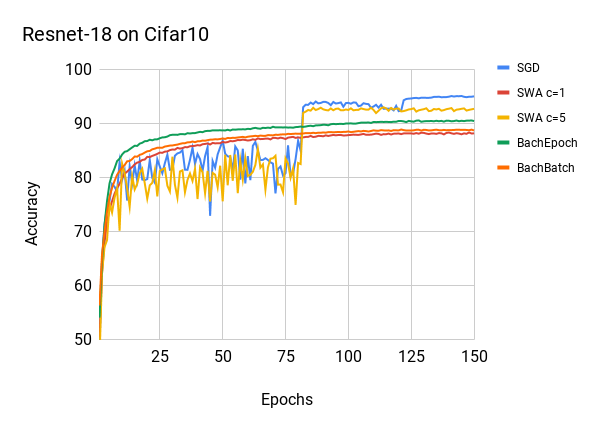}
\caption{Performance of aforementioned algorithms ~\cite{izmailov2018averaging,2012arXiv1212.2002L}. }
\label{fig:SWA}
\end{figure}

From \autoref{fig:SWA}, we see that none of the approaches can replicate the \ac{SGD} accuracy. This trend is consistent across different models under optimal hyperparameter settings. When using the \ac{SWA} technique in \cite{izmailov2018averaging}, the approaches deteriorate performance and hinder convergence. The performance improves with higher values of `c' because of fewer reassignments of the \ac{SWA} technique. \cite{2012arXiv1212.2002L} approach of using weighed averaging with more significance to later epochs in a linear (BachEpoch) and exponential (BachBatch) fashion also fail to converge optimally. All approaches add an increased computational load in processing of \ac{PRA} for each model parameter while training, reassignment of the computed values, followed by recalibration of \ac{BN} layers. These loads add up because all three tasks are performed for each epoch while training. Hence the techniques of ~\cite{izmailov2018averaging} and ~\cite{2012arXiv1212.2002L} which work impressively over improving generalization of pretrained neural networks and optimizing convex learning models respectively, when translated to \ac{DNN} training, increase the computational load and training time without providing optimal convergence.

\section{Methods}\label{sec:method}
\subsection{PSWA}

The analysis of the performance of the two methods shows that both \ac{SWA} and weighed averaging do provide better generalization at the early stages of the training process, typically in the underfitting regime. However, as the mean is biased by the model weights at the early stages of the training process, it cannot converge properly at the later stages of the training, even when one allows weighted averaging in favor of models at later training stages. We address this problem by removing the dependency of any prior weight distribution estimations for the general \ac{PRA} approach.

\begin{figure}[th]
\includegraphics[width=\columnwidth]{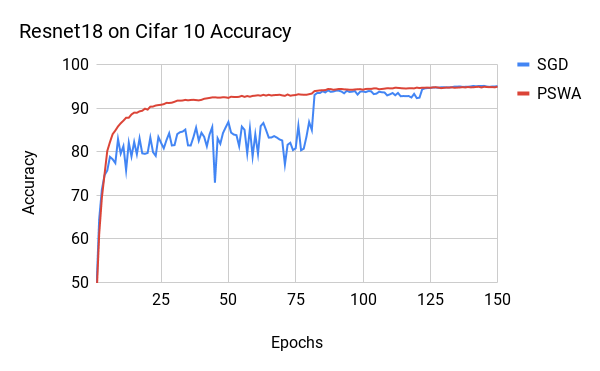}
\includegraphics[width=\columnwidth]{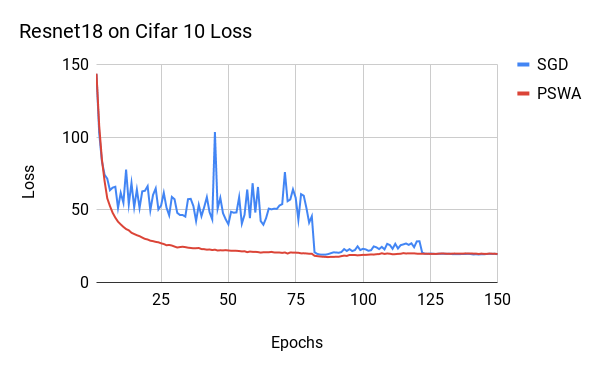}
\caption{Performance of PSWA on test data during training.}
\label{fig:r18}
\end{figure}

We call this technique \ac{PSWA}, as we sample the model weights over the batchwise updates, and repeat it periodically over each epoch. \autoref{fig:r18} depicts the application of \ac{PSWA} on ResNet18 for Cifar10 Accuracy and Cross Entropy Loss on the test dataset. We keep the experimental settings consistent with the prior experiments  as well as for the experiments in following section, where we simply use SGD to refer to SGD with momentum and L2 weight penalty and consistent learning rate schedules unless otherwise mentioned. The details of our implementation can be found in the Experiments section. The approach allows the model to train effectively for one epoch, while keeping running means for all model parameters over the weight distribution after batchwise gradient updates, followed by reassigning the running mean to the parameter weights, and then reinitializing the mean at the end of the epoch. This additional step allows for \ac{SGD} to gradually converge the model to the optimum by making gradient updates. Meanwhile, averaging over the batchwise distribution provides for a stabling effect on the model. 

An important challenge for applying general weight averaging techniques for \ac{DNN} models is the added computational load, which leads to longer training times.  
The time complexity of model training with weight averaging typically contains three parts:
  \begin{equation}\label{eq:time}
    T_\text{total} = T_\text{backprop} + \alpha T_\text{weight update} + \beta T_\text{calibrate BN},
  \end{equation}
where $T_\text{backprop}$, $T_\text{weight update}$, $T_\text{calibrate BN}$, mark the time spend on back-propagation, weight sampling, and Batch-Norm calibration using the full training dataset. Using the plain \ac{PSWA} for the same number of epochs clearly leads to a longer training time.

\begin{algorithm}[H]
\begin{algorithmic}[1]
\Require
\Statex Initialize \ac{DNN} model $\weight^{(0)}$
\Statex Initialize Learning rate schedule $\eta(e)$
\Statex Initialize training data batches $D_1$ ... $D_b$
\Statex Initialize total epochs $epochs$
\Statex Initialize running mean $\hat{\weight}$ for $\weight^{(0)}$ parameters
\Statex Determine sampling strategy and $\alpha,\beta$
\Ensure
\For{ $e$ in 1 ...  $epochs$}
	\State randomize($D_1$ ... $D_b$)
	\State reset ($\hat{\weight}$, 0)
	\For{ $i$ in 1 ...  $b$}
		\State $\weight^{(i)} = \weight^{(i-1)} - \eta\nabla_\weight l (\weight^{(i)}, D_i)$
		\State update ($\hat{\weight}$, $\weight^{(i)}$, $i$, $\alpha$) 
	\EndFor
	\State $\weight^{(e)}$ . assign($\hat{\weight}$)
	\State \ac{BN} recalibration ($\beta$)
\EndFor
\end{algorithmic}
\caption{\acl{PSWA}}
\label{algo:pswa}
\end{algorithm}

To remedy this additional computational load, we improve the plain-vanilla \ac{PSWA} such that we update the running mean for only a few percent ($\alpha$) of the randomly selected batches spread evenly over the training data; Similarly, we recalibrate the global mean and variance of each \ac{BN} layer with $\beta$ percent of the training data using a fast forward pass. We demonstrate later on that by reducing the number of updates in this fashion, the added computational cost becomes negligible for large datasets.

Algorithm 1 presents the general workflow of the \ac{PSWA} method for training a \ac{DNN}. After initializing the model parameters and data for training, we repeatedly update the model weights by \ac{SGD} or other gradient-based optimizations. Then we update the mean estimation of each weight.  The update is carried out in an online fashion. For $\alpha=100\%$ where we use the full dataset, it is:
\begin{equation}
\hat w \leftarrow \frac{i-1}{i} \hat w + \frac{w^{(i)}}{i} 
\end{equation}
To reduce the computational time, we only select $\alpha$ percent of batches to be used for mean estimation, and we change the count $i$ correspondingly. Before each epoch, we always reset the $\hat w$ to 0, and after the epoch, we reassign the mean weights to model weights.  After reassigning, the \ac{BN} layers are not best suitable for the new set of weights, so we recalibrate the \ac{BN} layers using $\beta$ percent of the training data to perform a forward pass and recompute global mean and variance statistics for each \ac{BN} layer.

\subsection{PWALKS and PSWM}

\begin{figure}
\centering
\includegraphics[width=\columnwidth]{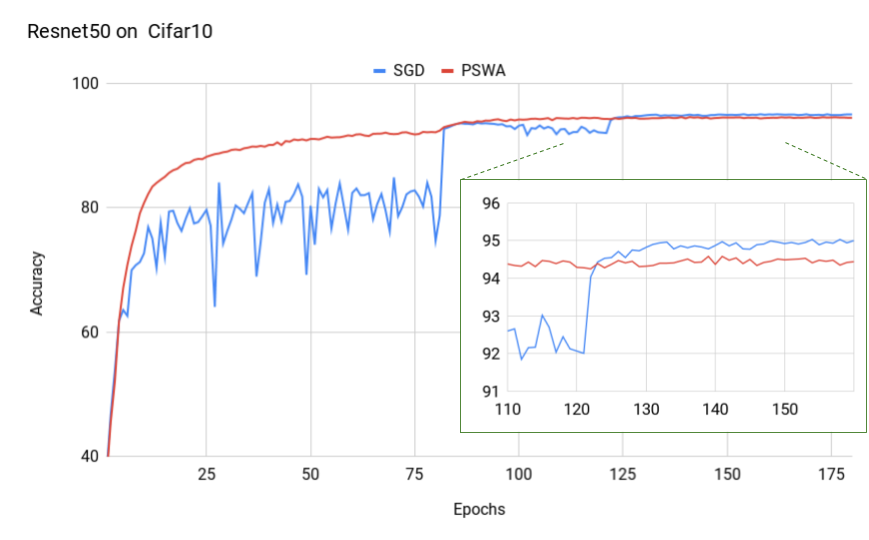}
\caption{PSWA on ResNet50 convergence problem.}
\label{fig:r50acc}
\end{figure}

Although the \ac{PSWA} method achieves optimal test accuracy using shallow ResNet18 model and other lightweight models, we found that for deeper networks \ac{PSWA} still does not converge properly to the optimum. \autoref{fig:r50acc} shows the effect on ResNet50. This problem is pervasive across similar deep networks like Inception and DenseNet, and also on datasets like ImageNet. However, it is important to note that the learning rate schedule decreases by a factor of 10 at epochs~80 and~120, and that it is only after the 120th epoch that the \ac{SGD} method converges to a better result than \ac{PSWA}. 

To investigate why this happens, we analyze \ac{PSWA} in more detail. In essence, for \ac{PSWA}, we modify the algorithm of \cite{2012arXiv1212.2002L} which works over the entire training process, to run over only one epoch without any prior. This approach does not burden the running mean with the weights distribution of the earlier epochs while still providing regularizing effect from \ac{PRA} over batchwise weight distribution. It however becomes cumbersome when the learning rate has decreased significantly as the batchwise descent of the \ac{SGD} loss function is able to reach a deeper minima for more complex models. We believe that by performing \ac{PRA} over the \ac{SGD} walk at this stage, the regularization counteracts the optimal convergence to the minima at the latter part of the batch-wise training. To address this problem we return to the conclusions presented by \cite{NIPS2011_4316}, who show that there is a need to carefully remove from the running mean, the initial weights which bias the mean towards the local minima.

We solve the suboptimal convergence problem for deeper networks by proposing two different modifications to \ac{PSWA}. In both approaches we allocate more importance to the model weights during the final batches while still maintaining the regularization afforded by using the weight distribution. 

\begin{figure}
\centering
\includegraphics[width=\linewidth]{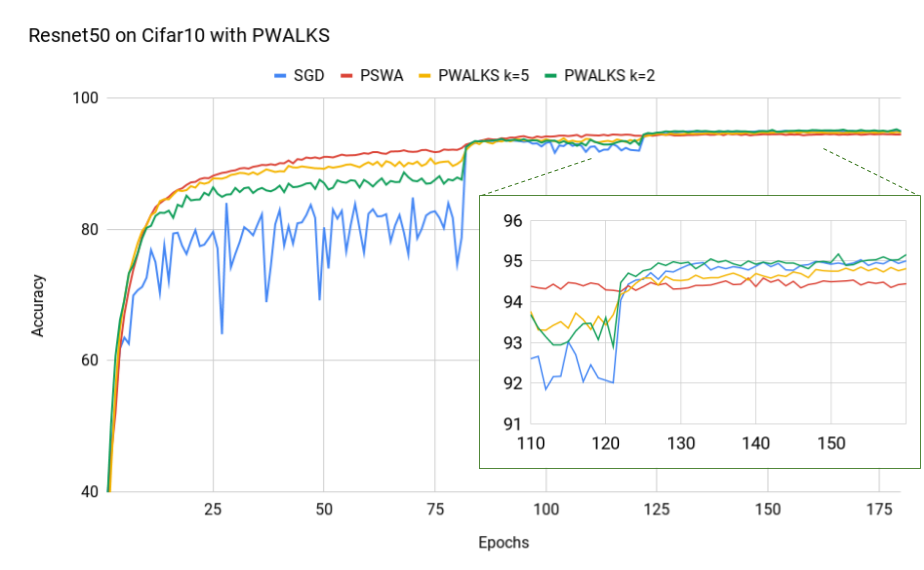}
\caption{PWALKS comparisons.}
\label{fig:pwalk}
\end{figure}

In the first approach, \ac{PWALKS}, instead of sampling weights evenly from all batches (for the mean weight distribution), we sample only the last `k'\% of the samples, `k' being a hyperparameter of size of the dataset and batches, ranging between 0 (last batch only, standard \ac{SGD}) and $100$ (\ac{PSWA} with $\alpha=1$). Empirically a small k value between 2-5 provides a consistently good performance by providing improvement over plain \ac{SGD} during early training (though not as much as \ac{PSWA}), and consistently converges to the optimum as demonstrated in \autoref{fig:pwalk}. Parallels can be drawn between the \ac{PWALKS} technique and constructing an ensemble of models over the last few batches, however, we incorporate the modified weights as part of the training process iteratively, unlike ensemble models. To convert \ac{PSWA} to \ac{PWALKS}, the update (line 7) of Algorithm \autoref{algo:pswa} is applied only when
\begin{equation}
 i > b \times (1-k\%).
\end{equation}
Moreover, the parameter k is equivalent to $\alpha$ in the \ac{PSWA} method in terms of controlling computational cost.

\begin{figure}
\centering
\includegraphics[width=\linewidth]{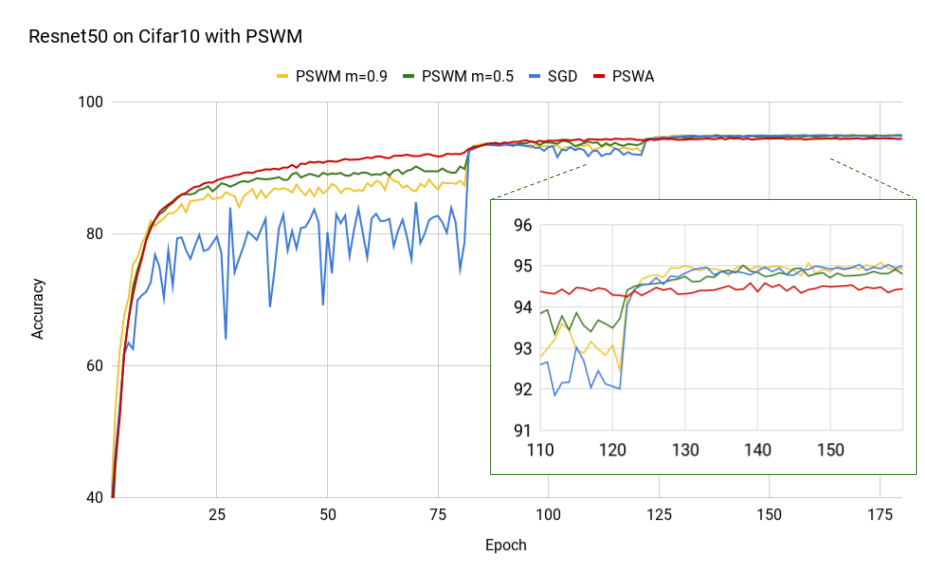}
\caption{PSWM comparisons.}
\label{fig:pswm}
\end{figure}

A second approach to solving the \ac{PSWA} convergence problem is to approach it from the perspective of a cumulative adjustment to weights. We propose a momentum-based modification to \ac{PSWA} called \ac{PSWM} where instead of keeping a running mean, we keep the running weights updated using momentum. For the model's parameters we keep a running momentum term, which we update at the end of each batch and reassign at the end of the epoch. Empirically momentum values between (0.5,0.9) yield good performance with $m =1$ being standard \ac{SGD}. To convert \ac{PSWA} to \ac{PSWM}, the update (line 7) of Algorithm \autoref{algo:pswa} is changed to:
\begin{equation}
\hat w \leftarrow (1-m) \times  \frac{i-1}{i} \hat w + m \times \frac{w^{(i)}}{i}.\end{equation}
And since the \ac{PSWM} is built on the \ac{PSWA}, the sampling technique developed for \ac{PSWA} can be applied to also reduce the time complexity of \ac{PSWM}.

\begin{figure}[th]
\includegraphics[width=\columnwidth]{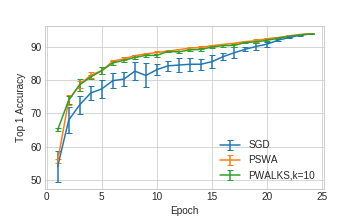}
\includegraphics[width=\columnwidth]{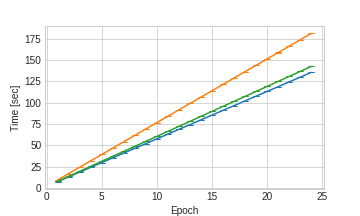}
\caption{Comparison of computation time for plain-vanilla \ac{SGD}, \ac{PSWA} with $\alpha=1$ and \ac{PWALKS}.}
\label{fig:performance}
\end{figure}

We next compare the computational performances of plain-vanilla \ac{SGD}, \ac{PSWA} with $\alpha=\beta=1$ and \ac{PWALKS} with $k=10$ and $\beta=10\%$.  The code is based on the fastest Cifar10 training code listed in the DAWN project (\cite{coleman2017dawnbench}); and the original implementation\footnote{Refer \url{https://github.com/davidcpage/cifar10-fast} for details (commit d31ad8d).} is changed from half-precision to full precision. We repeated the training process 10 times for each technique and report the corresponding mean and standard deviations. \autoref{fig:performance} shows that the \ac{PSWA} leads to a 34\% overhead when using the full training dataset for weight update and recalibration of \ac{BN} layers; and by adopting \ac{PWALKS}, we achieve the same prediction accuracy on the testing dataset without sacrificing the speed significantly without code-level optimizations, resulting in a 6\% overhead on training time. In addition, we also observe that the variations of the training process is much smaller when weight averaging techniques have been applied, which we discuss in results section.

\section{Experiments}\label{sec:experiments}

We have already covered experiments on Cifar10 with Resnet18 and Resnet50 using our techniques with SGD. The experimental settings of the previous experiments use Momentum values of 0.9, L2 weight penalty of 5e-4, learning rate decays of 0.1 at epochs of 80 and 120, for a total of 150 epochs. These experiments were performed with tuned hyperparameters and learning rate schedules to ensure optimal convergence. We now explore how our methods translate to other datasets, models and optimizers.

\subsection{Dataset: ImageNet}
\label{ssec:imag}
\begin{figure}[t]
    \includegraphics[width=\linewidth]{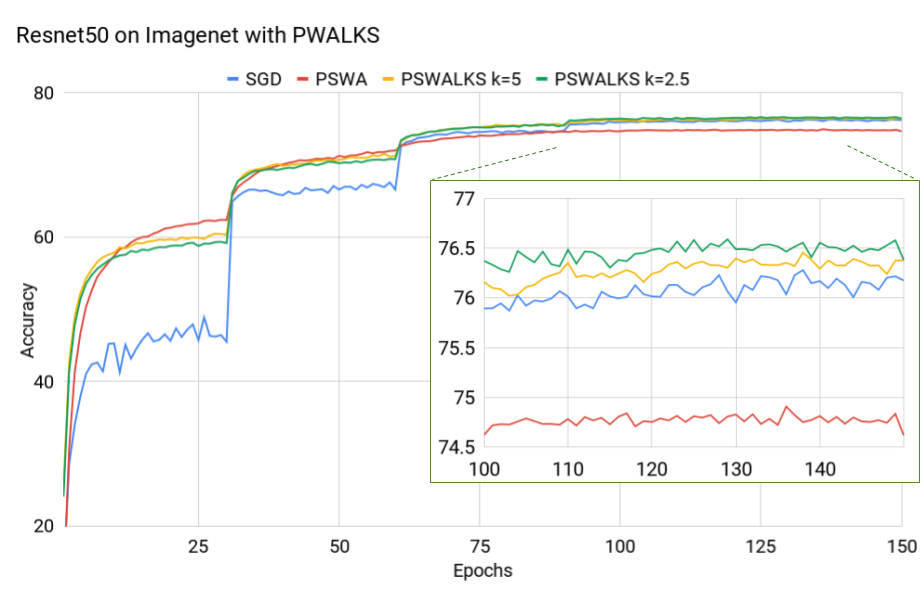}
\caption{ResNet50 on ImageNet with PWALKS.}
\label{fig:walk}
\end{figure}
\begin{figure}[t]
    \includegraphics[width=\linewidth]{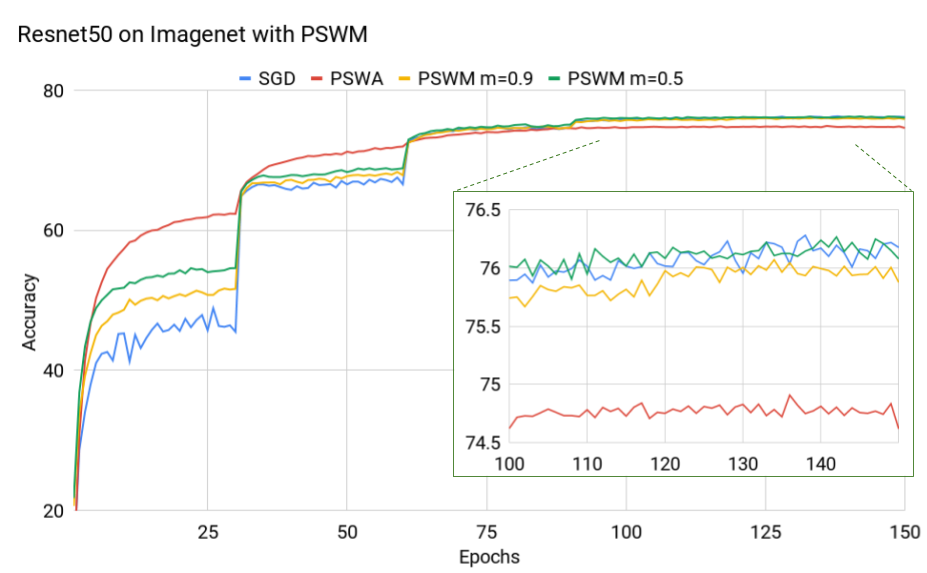}
\caption{ResNet50 on ImageNet with PSWM.}
\label{fig:swm}
\end{figure}

ImageNet~\cite{russ2015imagenet} is another standard image classification dataset, it has 1.2 million high resolution images from 1000 classes. Our implementations use ResNet50 as the underlying network, and SGD with momentum as the optimizer. We use learning rate with 0.1, which changes by a factor of 0.1 every 30 epochs, for a total of 150 epochs. Our results on ImageNet top-1 accuracy follow similar trends as on Cifar10 presented in \autoref{fig:walk} and \autoref{fig:swm} for \ac{PWALKS} and \ac{PSWM} respectively.

\subsection{Optimizer: Adam}
\label{ssec:adam}

\begin{figure}[t]
    \includegraphics[width=\linewidth]{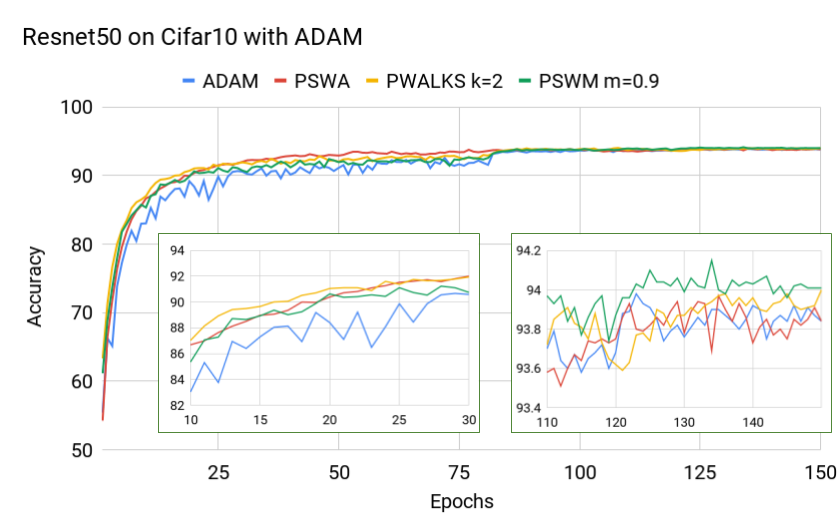}
    \caption{ResNet50 with Cifar10 using Adam.}
\label{fig:adam}
  \end{figure}

To show our techniques can be effectively used on adaptive optimizers as well, we present experiments on Adam ~\cite{kingma2015adam}, which performs first-order gradient-based optimization of stochastic objective functions, based on adaptive estimates of lower-order moments. \autoref{fig:adam} shows the implementation of ResNet50 on Cifar10 consistent with prior implementations except we use Adam instead of SGD, with a starting learning rate of 0.001. As we can see \ac{PSWA}, \ac{PWALKS} and \ac{PSWM} all offer marginal but consistent improvement on Adam, across epochs over multiple runs. The improvement is not as significant and dramatic as SGD, because Adam itself alleviates the common problems of SGD like large fluctuations, and slow convergence. Since Adam modulates the learning rate of each weight based on the magnitudes of its gradients, instead of the complete raw and noisy gradient vector, the distribution of the parameter weights remains small compared to SGD.

\begin{figure}[t] 
    \includegraphics[width=\linewidth]{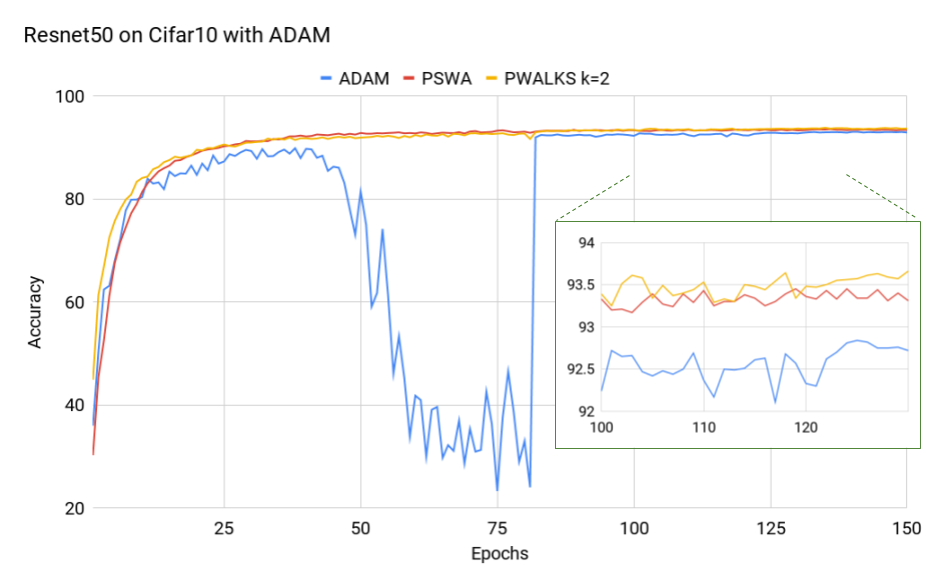}
\caption{ResNet50 on Cifar10 with Adam and high learning rate.}
\label{fig:adamp}
\end{figure}

Adam and other adaptive optimizers suffer from some important documented problems. Though Adam converges faster, it does not generalize well (\cite{keskar2017improving}). From our experiments \ac{PSWA} over Adam also provided for reduced CrossEntropy loss over training. Another problem for adaptive optimizers like RMSPROP and Adam is they become unstable at high learning rates near convergence. This happens as the squares of rolling mean of gradients are used to divide the current gradient, in which case very small gradients can introduce instability. In \autoref{fig:adamp} we present such a scenario where we use a $l$ of 0.01 (instead of 0.001) which causes Adam to become unstable. However, Adam with \ac{PSWA} remains stable and converges better.

\begin{figure*}
\centering
\begin{tabular}{ccc}
\includegraphics[width=5.5cm]{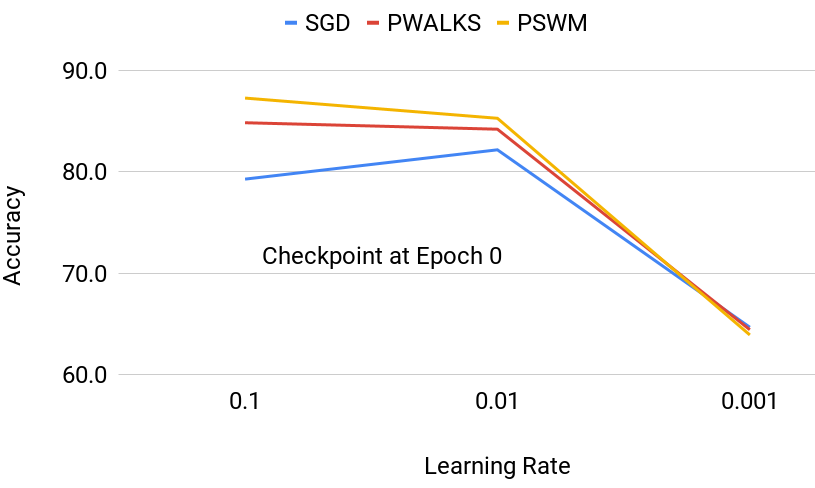} &
\includegraphics[width=5.5cm]{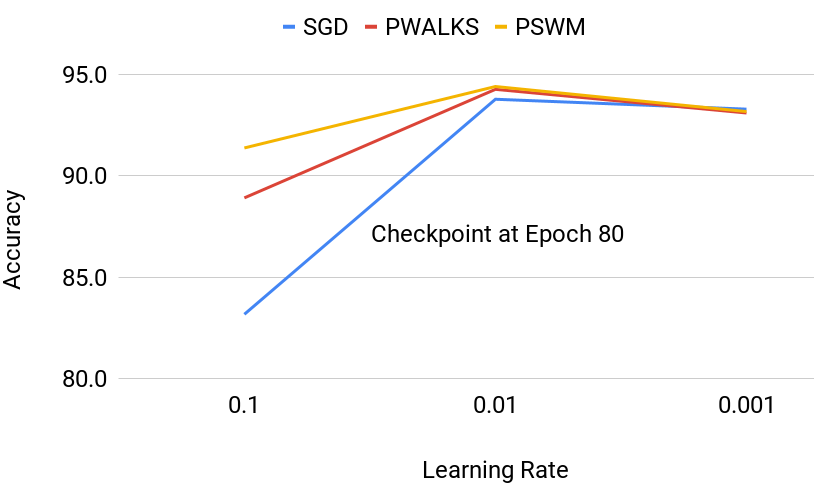} &
\includegraphics[width=5.5cm]{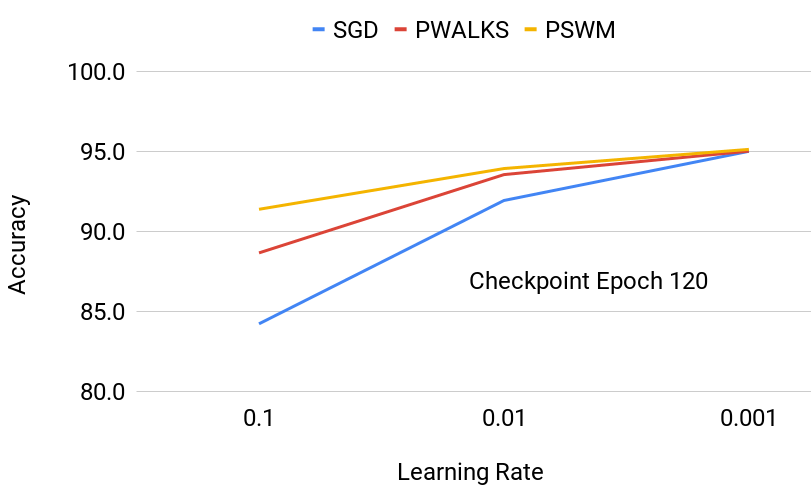} \\
\end{tabular}
    \caption{Faster convergence from different checkpoints and learning rates on Cifar10}
    \label{fig:checkpoint} 
\end{figure*}

\subsection{Other Experiments}
We perform more experiments using different model architectures like MobileNet  \cite{sandler2018mobilenetv2}, Inception \cite{szegedy2016rethinking}, Densenet \cite{huang2017densely} which are presented in the supplementary paper along with experiments on Human Pose detection and Scene segmentation to cover a broad array of Computer Vision applications. We perform extensive experiments on Cifar10 under different configurations such as suboptimal learning rates and retraining from checkpoints, since these experiments are faster and their results generalize well to larger datasets like Imagenet. These experiments help analyze the strengths of our algorithms discussed in Results section.

\section{Results}\label{sec:analysis}

In this section we analyze the performance of our algorithms and try to quantify the improvements afforded by them. We use performance statistics of Resnet18 and Resnet50 on Cifar10 and Imagenet respectively, and compare PSWA, PWALKS (k=2), PSWM (m=0.5) against the baseline implementations which use \ac{SGD}. We see for both Cifar10 and Imagenet, \ac{PSWA} converges much faster during early training, but does not converge optimally. \ac{PWALKS} provides better generalization than \ac{PSWM}, while both the techniques provide improvement over SGD. Hence in this section we focus more on \ac{PWALKS}, owing to its superior empirical performance. We show how our methods can provide faster, better and more robust performance empirically across different datasets, models and experimental settings. We note that since these terms have an overlap in their definitions, it is important how we interpret them in the the context of non-convex optimization and deep learning. Moreover, it is non-trivial to quantify this improvement as both the accuracy and loss functions form a non-stationary and volatile time-series.

\subsection{Faster}

We can interpret an algorithm being faster than a baseline approach if it achieves a performance threshold before the baseline. For our results we use validation thresholds of 90\%, 95\% for Cifar10; 92\%, 93\% \cite{coleman2017dawnbench} for top-5 and 75\% for top-1 for Imagenet. These thresholds represent performance at near convergence and at convergence for the corresponding models and datasets, we present the results in \autoref{tab:threshold}. It is important to note that for Imagenet, \ac{PSWM} and \ac{PWALKS} cross the at convergence thresholds not only with fewer epochs compared to the baseline approach but with larger learning rates. SGD needs learning rate to decrease by a factor 0.1, under the same experimental settings to cross the thresholds.

\begin{table}[t]
  \caption{Epochs to achieve various accuracy thresholds.}
  \label{tab:threshold}
  \begin{tabular}{ccccc}
  \thead{} & \thead{ Baseline } & \thead{ PSWA } & \thead{ PSWM } & \thead{ PWALKS }\\
    Cifar10, 90\% & 81 & 22 & 22 & 64\\
    Cifar10, 95\% & 137 & N.A. & 128 & 140\\
    Imagenet (top 1) 75\% & 91 & N.A. & 72 & 80 \\
    Imagenet (top 5) 92\% & 69	& 83 & 66 & 65\\
    Imagenet (top 5) 93\% & 129 & N.A. & 103 & 93\\
\end{tabular}
\end{table} 

Another interpretation of faster can be in terms of convergence of the optimization algorithm for various stages and learning rates of the training process. We train Resnet18 on Cifar10 using the baseline SGD implementation and save checkpoints at epochs 0, 80, 120. We then retrain the checkpoints under different learning rates for 10 epochs and compare the results in \autoref{fig:checkpoint}. As we can see, for all starting checkpoints, our algorithms converge much faster than baseline SGD at high and optimal learning rates. When the learning rate is very small, variances of the weight distributions are correspondingly small as well and hence our algorithms match the SGD performance.  

\subsection{Better}

We emphasize that in all our presented examples, the optimizer, learning rate (and its schedule), and training hyperparameters, have all been finetuned for convergence for the original baseline implementations, and we do not modify them when using our techniques to ensure fair comparisons. But for most machine learning applications we are unaware of these optimal hyperparameters and learning rate schedules which can introduce volatility in the training process and uncertainty regarding the final convergence. We simulate these conditions on Cifar10 with Resnet18, while keeping the same experimental setting as before but with initial learning rates which are suboptimal. The results are presented in \autoref{tab:suboptimal}. As we can see when we have suboptimal training routines, we can still expect better performance and generalization from \ac{PSWM} and \ac{PWALKS}. 

\begin{table}[t]
  \caption{Accuracy when training Resnet18 on Cifar10 with initial suboptimal learning rates. Optimal learning rate is 0.1}
  \label{tab:suboptimal}
  \begin{tabular}{ccccc}
  \thead{ Algorithm } & \thead{ LR = 1 } & \thead{ LR = 0.5 } & \thead{ LR = 0.05 } & \thead{ LR = 0.01 }\\
SGD & 85.2 & 91.2 & 94.9 & 94.1 \\
PWALKS & 86.6 & 91.6 & 94.9 & 94.1 \\
PSWM & 88.0 & 91.6 & 95.2 & 94.1 \\
\end{tabular}
\end{table}

\setlength{\tabcolsep}{1em}
\begin{table}[t]
  \caption{Intermediate accuracy improvement when training ResNet50 on ImageNet.}
  \label{tab:inter_improv}
  \begin{tabular}{ccc}
   & \thead{ Average Improvement \\over SGD accuracy \\ for all epochs} &	\thead{ Average Improvement \\ over SGD accuracy \\for first 30 epochs }\\
    PSWA & 2.75 &	12.52\\
    PWALKS & 3.50 &	12.37 \\
    PSWM & 2.01 &	7.81 \\
\end{tabular}
\end{table}

Since the baseline models are tuned to reach convergence, we can also analyze the intermediate performance during the training process under optimal experiment settings. We focus on intermediate performance both over the entire training process and during the early stages at high learning rates, presented in \autoref{tab:inter_improv}. These scenarios have direct applications in online learning and learning on resource constrained devices respectively, as presented in Discussions section. From these results it is clear that when we are dealing with suboptimal training settings our algorithms can provide better performance, and under optimal training settings we still provide better intermediate performance. We can also interpret \autoref{fig:checkpoint} in terms of better performance under different initial settings and learning rates.

\subsection{Robust}

We analyze the test accuracy distribution with ResNet50 over ImageNet and discover all three of our techniques provide consistent and significantly lower \ac{SD}, both at convergence and over the saturation phases at constant learning rates as shown in \autoref{tab:freq}. The lower variance at early saturation phase (epoch 20-30) points to a less volatile training process and the lower variance at the final 20 epochs point to a more stable convergence. Another important observation is that \ac{PWALKS} and \ac{PSWA} performance on the test set monotonically increases or remains stable over epochs until convergence. This is especially important since it indicates that with a high probability the model is consistently improving and the performance does not sporadically fluctuate like the baseline model. Again analyzing test accuracy distribution with ResNet50 over ImageNet, we find that for almost 70\% consecutive epochs with \ac{PSWA} the accuracy is improving, or 99\% of them are stable within 0.2 percentage decrement, unlike \ac{SGD}-based training which only shows 57\% and 86\% respectively (shown in \autoref{tab:improv}). We also compare performance improvements between current epoch and best overall performance over previous epochs and present the stability within 0.2 percentage decrement. As we can see PSWA improves upon best previous performance or remains stable for 99\% of the epochs.

\setlength{\tabcolsep}{0.5em}
\begin{table}[t]
  \caption{Volatility analysis of training ResNet50 on ImageNet.}
  \label{tab:freq}
  \begin{tabular}{cccc}
   & \thead{Standard Deviation \\ last 20 epochs} &	\thead{Standard Deviation \\ epochs 20-30}	& \thead{Mean \\ last 20 epochs}\\
SGD	         & 0.080 & 1.06	& 76.16\\
PSWA         & 0.058 & 0.39 & 74.77\\
PWALKS & 0.047 & 0.21 & 76.50\\
PSWM  & 0.056 & 0.24 & 76.16\\
\end{tabular}
\end{table}

\begin{table}[t]
  \caption{Monotonic improvement demonstrating stability when training ResNet50 on ImageNet. (\% of epochs )}
  \label{tab:improv}
  \begin{tabular}{cccc}
   & \thead{ improvement \\ over previous } &	\thead{ stable improvement \\ over previous  } &	\thead{ stable improvement \\ over max previous }\\
    SGD           & 57\% & 86\% & 77\% \\
    PSWA          & 70\% & 99\% & 99\% \\
    PWALKS      & 65\% & 98\% & 97\% \\
    PSWM    & 65\% & 96\% & 90\% \\
\end{tabular}
\end{table}

\section{Discussions and Additional Details}

Following the Results section we see these advantages can significantly improve several computer vision based applications. Since \ac{PWALKS} performs better at high learning rates (which offers training speed-up) and plateaus over shorter time, it can be especially useful when training on large datasets with constrained computational resources that require training at high learning rates for short time periods. These conditions arise for deep learning on Edge Devices where the models cannot realistically be trained to convergence. Moreover, \ac{PWALKS}' stability and monotonic improvements during training can uniquely benefit deployed models learning in an online fashion or over continuous streaming data, since \ac{PWALKS} ensures models will not sporadically deteriorate during different phases of the training process. 

Our analysis of the loss surface (presented in the supplementary paper) shows that these techniques produce minima that are deeper and wider than those found by \acs{SGD}. The details of the implementations and system information necessary for reproducibility are also provided in supplementary paper along with other experiments. Also, while both \ac{PWALKS} and \ac{PSWM} offer optimal convergence, they depend on hyperparameters `k' and `m' respectively. Empirically we find that both techniques converge optimally to within a small margin to each other across a wide range of their respective hyperparameter values, and there isn’t an explicit need for researchers to tune them.

\section{Conclusion}\label{sec:conclusion}

In this paper, we introduced a trio of techniques (PSWA, PWALKS, and PSWM) based on sampling over model weights that solve issues with previous weight averaging approaches. Our proposed techniques provide increased robustness and stability to the training process (\autoref{tab:improv}) of neural networks as well as substantial intermediate performance improvements (\autoref{tab:inter_improv}) without compromising on optimal convergence (\autoref{tab:freq}). PWALKS’ almost monotonic and stable performance (\autoref{tab:improv}) ensures performance does not fluctuate depending on the current batch. As (\autoref{tab:threshold}) reflects, our techniques can provide faster convergence for thresholds and for different checkpoints under diverse experimental settings (\autoref{fig:checkpoint}). In the light of the advantages offered by these techniques, they provide a good starting point for training DNNs especially in those cases where no optimal training regime exists. 

\section*{Supplementary Material}\label{sec:supp}

\begin{table*}[h]
\centering
\begin{tabular}{l|c|c}
\hline
&\textbf{Stochastic Gradient Descent} & \textbf{PSWA (with SGD)} \\
\hline \hline 
& \underline{Initialize:} \quad $\pmb{w}^{1,0}\leftarrow 0$ & $\pmb{w}^{1,0}\leftarrow 0$ \\
\multirow{4}{*}{\rotatebox{90}{\textbf{Epoch 1}}} & $\begin{aligned}[t] 
\pmb{w}^{1,1}\leftarrow \pmb{w}^{1,0} - \eta \times \pmb{g}^{1,0}\\
\pmb{w}^{1,2}\leftarrow \pmb{w}^{1,1} - \eta \times \pmb{g}^{1,1}\\
\vdots \\
\pmb{w}^{1,T}\leftarrow \pmb{w}^{1,T-1} - \eta \times \pmb{g}^{1,T-1}
\end{aligned}$ &
$\begin{aligned}[t] 
\pmb{w}^{1,1}\leftarrow \pmb{w}^{1,0} - \eta \times \pmb{g}^{1,0}\\
\pmb{w}^{1,2}\leftarrow \pmb{w}^{1,1} - \eta \times \pmb{g}^{1,1}\\
\vdots \\
\pmb{w}^{1,T}\leftarrow \pmb{w}^{1,T-1} - \eta \times \pmb{g}^{1,T-1}\\
\end{aligned}$ \\
\hline 
& \underline{Initialize:} \quad $\pmb{w}^{2,0} \leftarrow \pmb{w}^{1,T}$ & $ \pmb{w}^{2,0} \leftarrow \frac{1}{T} \sum_{t=1}^{t=T} \pmb{w}^{1,t}$ \\
\multirow{4}{*}{\rotatebox{90}{\textbf{Epoch 2}}} &$\begin{aligned}[t] 
\pmb{w}^{2,1}\leftarrow \pmb{w}^{2,0} - \eta \times \pmb{g}^{2,0}\\
\pmb{w}^{2,2}\leftarrow \pmb{w}^{2,1} - \eta \times \pmb{g}^{2,1}\\
\vdots \\
\pmb{w}^{2,T}\leftarrow \pmb{w}^{2,T-1} - \eta \times \pmb{g}^{2,T-1}\\
\end{aligned}$ &
$\begin{aligned}[t] 
\pmb{w}^{2,1}\leftarrow \pmb{w}^{2,0} - \eta \times \pmb{g}^{2,0}\\
\pmb{w}^{2,2}\leftarrow \pmb{w}^{2,1} - \eta \times \pmb{g}^{2,1}\\
\vdots \\
\pmb{w}^{2,T}\leftarrow \pmb{w}^{2,T-1} - \eta \times \pmb{g}^{2,T-1}\\
\end{aligned}$ \\
\hline
& \vdots & \vdots \\
\hline
\multirow{6}{*}{\rotatebox{90}{\textbf{Epoch E}}} & $\begin{aligned}[t] 
\text{\underline{Initialize:}} \quad \pmb{w}^{E,0} \leftarrow \pmb{w}^{E-1,T} \\
\pmb{w}^{E,1}\leftarrow \pmb{w}^{E,0} - \eta \times \pmb{g}^{E,0}\\
\pmb{w}^{E,2}\leftarrow \pmb{w}^{E,1} - \eta \times \pmb{g}^{E,1}\\
\vdots \\
\pmb{w}^{E,T}\leftarrow \pmb{w}^{E,T-1} - \eta \times \pmb{g}^{E,T-1}\\
\pmb{w}^{Final} \leftarrow \pmb{w}^{E,T} \\
\end{aligned}$ &
$\begin{aligned}[t] 
\pmb{w}^{E,0} \leftarrow \frac{1}{T} \sum_{t=1}^{t=T} \pmb{w}^{E-1,t} \\
\pmb{w}^{E,1}\leftarrow \pmb{w}^{E,0} - \eta \times \pmb{g}^{E,0}\\
\pmb{w}^{E,2}\leftarrow \pmb{w}^{E,1} - \eta \times \pmb{g}^{E,1}\\
\vdots \\
\pmb{w}^{E,T}\leftarrow \pmb{w}^{E,T-1} - \eta \times \pmb{g}^{E,T-1}\\
\pmb{w}^{Final} \leftarrow \frac{1}{T} \sum_{t=1}^{t=T} \pmb{w}^{E,t} \\
\end{aligned}$ \\
\hline
\end{tabular}
\caption{Standard SGD and \ac{PSWA} algorithm updates for E epochs (with T iterations per epoch).}
\label{tab:pswa_sgd} 
\end{table*}

\section{Motivation}\label{sec:Motivation}
Our motivation for intermediate averaging stems from the stability induced into the standard \ac{SGD}  updates through various averaging schemes \cite{shamir2013stochastic}. We illustrate this by comparing the standard SGD with the periodically sampled weight averaging (PSWA). The layout of the updates for $E$ epochs (with $T$ iterations per epoch) for the algorithms is provided in \autoref{tab:pswa_sgd}. Here we define, $\pmb{w}^{e,t}$ =  model weights during the $e^{th}$ epoch and $t^{th}$ iteration within an epoch, with $t = 0 \ldots T$ and, $e = 0 \ldots E$ \footnote{$0$ stands for the initial values}. For simplicity we fix the batch size for SGD updates to 1.  Further, $\eta$ is the constant learning rate, and at each step an oracle provides with the vector $\mathbb{E}(\pmb{g}^{e,t}) \in \partial f(\pmb{w}^{e,t})$, where $\mathop{\mathbb{E}}$ is the expectation operator. Finally, we use $\pmb{w}^\ast$ as the model weights at convergence. 

Next, we define a notion of stability which simply captures the variance of the expected function values at the end of each epoch from convergence. 

\begin{mydef} Stability of an algorithm $\mathcal{A}$ is defined as
\begin{equation*}
  \mathcal{S}_{\mathcal{A}} = \| \pmb{s}^{\mathcal{A}} \| ^{2} _{2}
\end{equation*}
where \\
\begin{equation*}
  \pmb{s}^{\mathcal{A}} = [ \, \pmb{s}^{\mathcal{A}} _{1} ...  \pmb{s}^{\mathcal{A}} _{e} ... \pmb{s}^{\mathcal{A}} _{E} ] \; \forall e = 1 \ldots E 
\end{equation*}
and 
\begin{equation*}
  \pmb{s}_{e}^{\mathcal{A}} = \mathop{\mathbb{E}} ( f ( \pmb{w}_{e}^{\mathcal{A}} ) -   f ( \pmb{w}^{\ast} ) ) 
\end{equation*}
\end{mydef}
From this definition it is clear that it is desirable to have smaller $\mathcal{S}_{\mathcal{A}}$ for more stable algorithms.

Next, to analyze the stability of the standard SGD and the PSWA method, we use the Theorems \autoref{thm:sgd} and \autoref{thm:sgd_avg} respectively, presented next. Theorem  \autoref{thm:sgd}  provide the best provable convergence rate for standard SGD with the following assumptions,

\begin{assumption} \label{assume} We make the minimal assumptions,
\begin{enumerate} 
    \item[\textbf{(A1)}] The function $f$ is convex in $\pmb{w}$ with convex domain $\mathcal{W}$.
    \item[\textbf{(A2)}] Bounded difference i.e. there exists a constant  $D = \underset{\pmb{w},\pmb{w}' \in \mathcal{W}}{sup} ||\pmb{w} - \pmb{w}'||_2^2$ with $D< \infty$.
    \item[\textbf{(A3)}] Bounded gradients i.e.  $\mathbb{E}(||\mathbf{g}^{e,t})||_2^2 \leq G^2$. 
\end{enumerate}
\end{assumption}

Note that here although our analysis assumes a convex function, we make mild assumptions on the nature (stability) of the functions. Under Assumption \autoref{assume} we have the following,

\begin{mytheorem} \label{thm:sgd} Under Assumption \autoref{assume} the standard SGD algorithm after $e$ epochs (with $T$ iterations per epoch) with a step size $\eta_t = 1/\sqrt{t}$, \; $t = 1 \ldots Te$ provides, 
\begin{flalign}
&\mathop{\mathbb{E}} [ \, f ( \pmb{w}_{e,0}^{SGD}) - f ( \pmb{w}^{\ast} ) ] \leq (D^2 + G^2) \frac{2+log(T(e-1))}{\sqrt{T(e-1)}}  \nonumber && \\ 
&  \quad \quad \quad  \quad \quad \quad  \quad \quad \quad \quad  \simeq \; \mathbb{O} ( \, \frac{\log (Te)}{\sqrt{Te}} ) \nonumber &&
\end{flalign}
\end{mytheorem}
\noindent \textbf{Proof}: The proof follows by direct application of Theorem 2 in \cite{shamir2013stochastic} for T(e-1) iterations. \qed 

\noindent In fact, under the similar assumptions we see that the PSWA algorithm follows,

\begin{mytheorem}\label{thm:sgd_avg}
Under Assumption \autoref{assume} the PSWA algorithm after $E$ epochs (with $T$ iterations per epoch) and a constant step size  $\eta = \sqrt{\frac{D^2}{T G^2}}$ provides,
\begin{align}
&\mathop{\mathbb{E}} [ \, f ( \pmb{w}_{e,0}^{PSWA}) - f ( \pmb{w}^{\ast} ) ] \leq \frac{(DG)}{\sqrt{T}}  \simeq \; \mathbb{O} ( \, \frac{1}{\sqrt{T}} ) \nonumber &&
\end{align} 
\end{mytheorem}
\noindent \textbf{Proof}: The proof follows through applying Theorem 14.8 in \cite{shalev2014understanding} for the final epoch $E-1$ of PSWA and bounding $||\pmb{w}^{e-1,0} - \pmb{w}^*||_2^2 \leq D^2$. \qed

\noindent \textbf{Remark} A straight-forward comparison from Theorems 1 \& 2 and using the Definition 1 shows that, for sufficiently large $T \gg e$, the PSWA has better stability i.e. $\pmb{s}_{e}^\text{PSWA} \simeq \mathbb{O} (\, \frac{ 1 }{\sqrt{T}} ) \leq  \mathbb{O} ( \, \frac{\log (Te)}{\sqrt{Te}}) \simeq \pmb{s}_{e}^\text{SGD}$. This indicates through periodically sampling we can improve the stability of the algorithm at least in the initial stages of the algorithm where $T \gg e$. This intuition is further validated through our Experiments in Section 3 of the main paper.

\section{Additional Experiments}\label{sec:add_exp}

\subsection{Dataset: Cifar10}
\label{ssec:cif}

We have already discussed in detail the results of our techniques on ResNet18 and ResNet50 over Cifar10. ResNet18 is trained for 150 epochs and ResNet50 is trained for 180 epochs. Both use \ac{SGD} with momentum of 0.9, L2 penalty of 0.0005, and have a learning rate schedule which decreases by factor of 10 at epochs 80,120 and 150.  We use Standard CrossEntropy loss and batch size of 128.
\begin{figure}[h]
    \includegraphics[width=\linewidth]{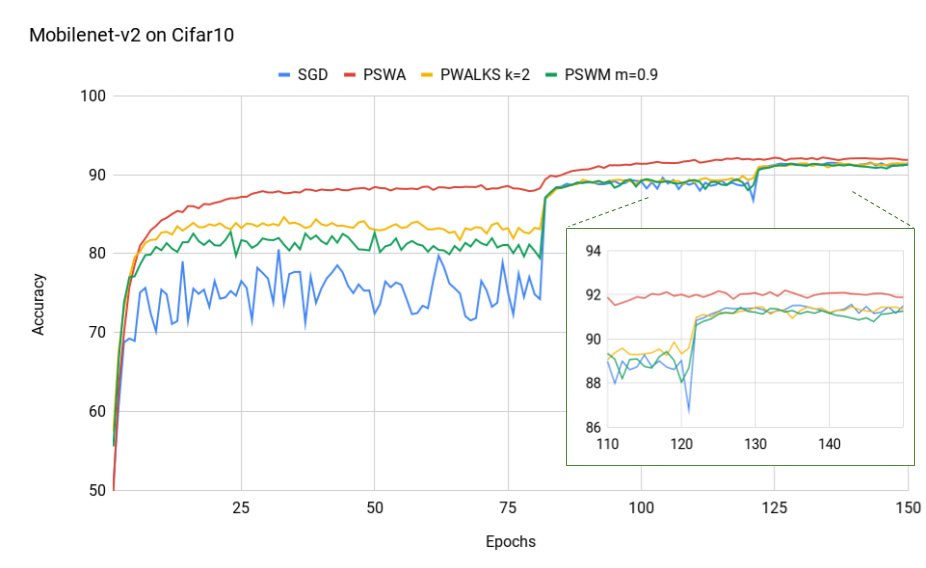}
\caption{MobileNet trained on Cifar10.}
\label{fig:mobilenet}
\end{figure}
\begin{figure}[h]
    \includegraphics[width=\linewidth]{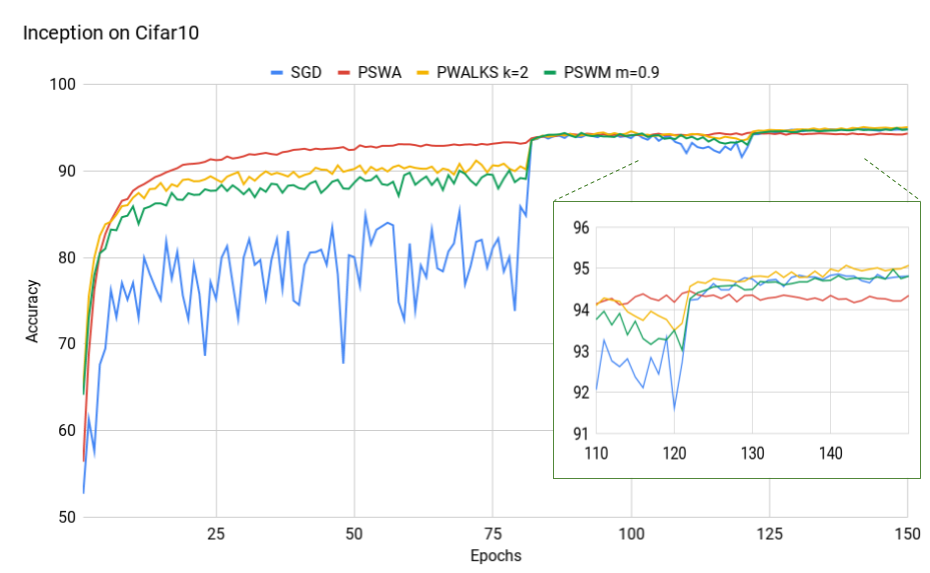}
    \caption{Inception trained on Cifar10.}
\label{fig:inception}
\end{figure}

In our experiments on shallow networks like MobileNet-v2 \cite{sandler2018mobilenetv2} and ResNet18, we find \ac{PSWA} not only provides faster and more robust convergence, but also converges to a more optimal minima, as evident in \autoref{fig:mobilenet}. Another interesting comparison is between \ac{PWALKS} and \ac{PSWM} with \ac{PSWA} on shallow networks, where \ac{PSWA} converges to deeper minima, which \ac{PWALKS} and \ac{PSWM} are unable to. However, for deeper networks like Inception~\cite{szegedy2016rethinking} , DenseNet-121~\cite{huang2017densely} and ResNet50, as discussed before, \ac{PSWA} does not converge properly, while both \ac{PWALKS} and \ac{PSWM} do. \autoref{fig:inception} shows Inception network trained using the same implementation as above. We observe that \ac{PSWA} and its variations reach 90\% and 94\% thresholds much faster consistently and while training on larger learning rate, while SGD needs a learning rate change by a factor of 10, to cross the thresholds. 

\subsection{Task: Human-Pose Detection}
\label{ssec:hpii}

We apply our techniques on the work of \cite{xiao2018simple}, where they perform 
Human-keypoint detection on MS-COCO \cite{lin2014microsoft} and Human-pose detection on MPII dataset \cite{andriluka14cvpr} \footnote{An aberrant drop in PSWA accuracy in Figure 15, seems a result of biased subset of data points during \ac{BN} recaliberation.}. Both tasks use ResNet50 pretrained on ImageNet, and perform transfer learning on the new dataset. Both experiments use Adam as the optimizer with a learning rate of 0.001. Consistent with our prior experiments, \ac{PWALKS} and \ac{PSWM} provide consistent improvement over Adam in the early stages of training.

\begin{figure}[h]
    \includegraphics[width=\linewidth]{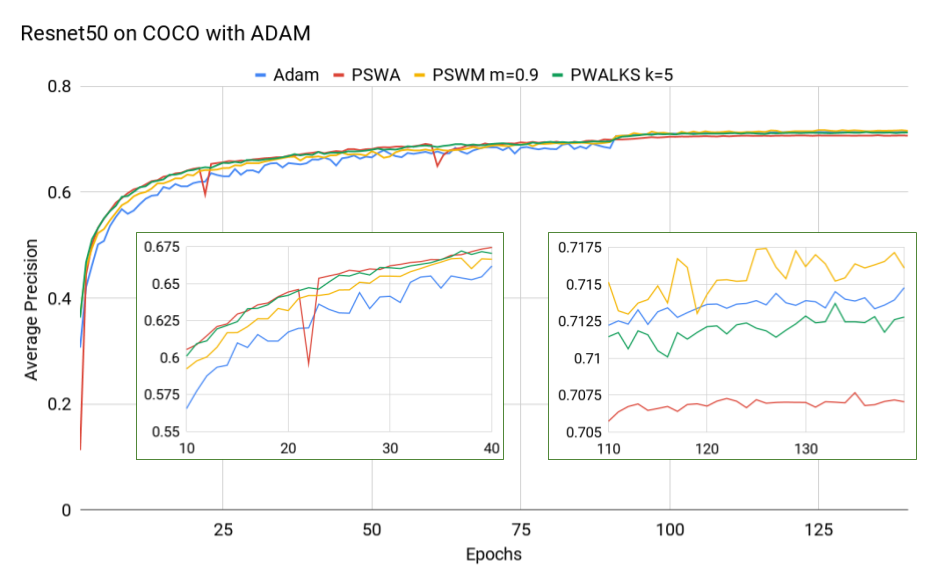}
\caption[Object detection]{COCO-Keypoint detection on ResNet50 and Adam}
\label{fig:coco}
\end{figure}

\begin{figure}[h]
    \includegraphics[width=\linewidth]{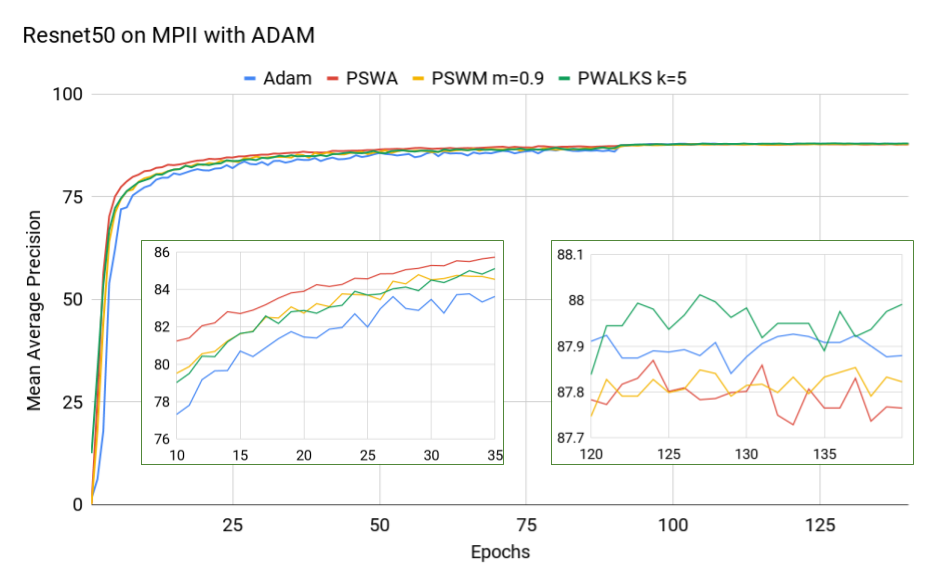}
\caption{MPII Human-Pose detection on ResNet50 and Adam}
\label{fig:mpii}
\end{figure}

\subsection{Task: Segmentation}
\label{ssec:seg}

We also apply our techniques on the works of \cite{zhou2017scene}, where they perform scene segmentation on MIT ADE20K Dataset \cite{zhou2016semantic}, the largest open source dataset for semantic segmentation and scene parsing. The implementation uses an encoder-decoder architecture with ResNet50 pretrained on ImageNet as the encoder and Pyramid Pooling Module with Bilinear Upsampling as decoder with deep supervision \cite{xiao2018unified}. The implementation uses per-pixel cross-entropy loss, SGD as the optimizer and a 'poly' learning rate policy.

\begin{figure}[h]
    \includegraphics[width=\linewidth]{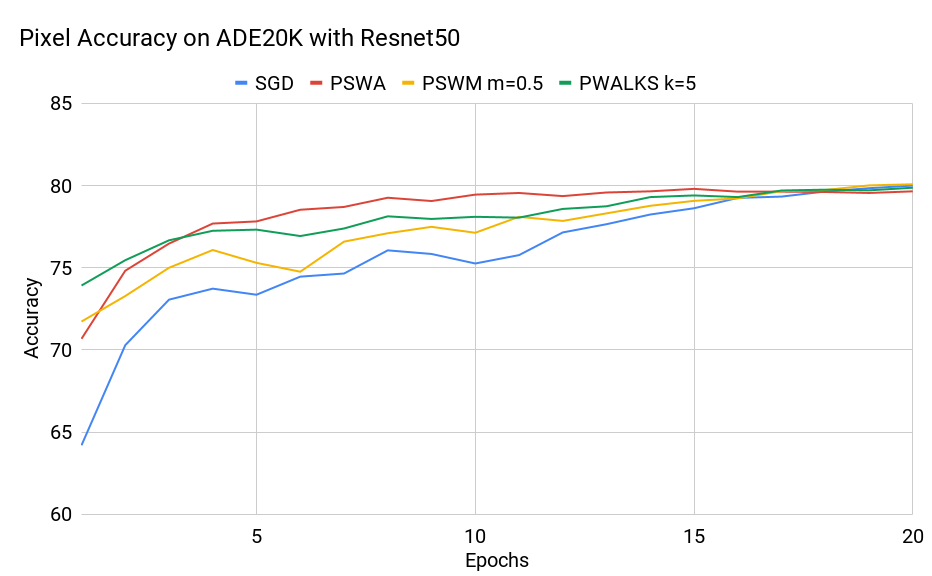}
\caption{Pixel accuracy of segmentation on ADE20K.}
\label{fig:pixel}
\end{figure}

\begin{figure}[h]
    \includegraphics[width=\linewidth]{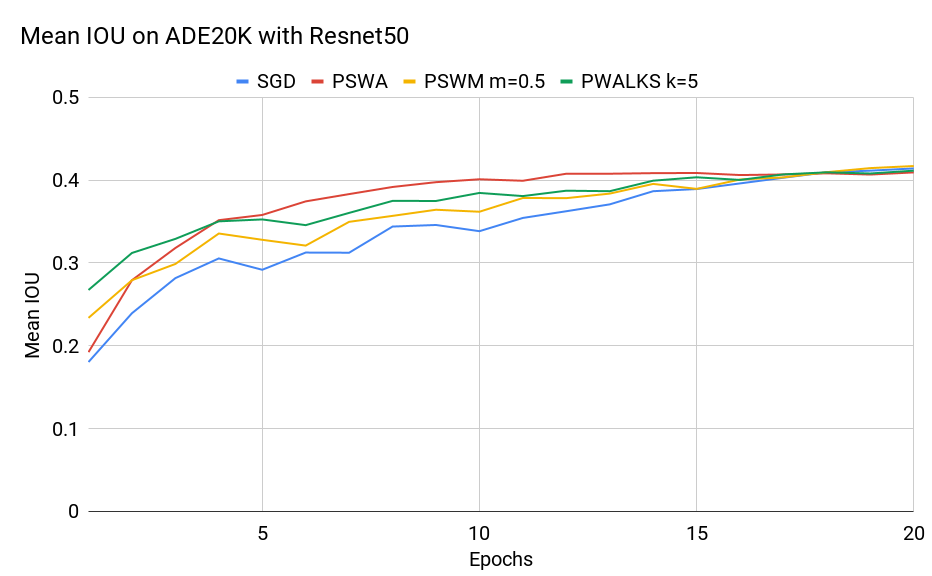}
\caption{Mean IOU of segmentation on ADE20K.}
\label{fig:mou}
\end{figure}

For our implementation we initialize two distributions one each for the encoder and decoder. We update both the distributions together and reassign at the end of the epoch. We do not need to recalibrate the \ac{BN} layers, since the implementation uses Synchronized Batch Normalization \cite{peng2017megdet}. \autoref{fig:pixel} shows Pixel wise accuracy of the Segmentation models on test set and \autoref{fig:mou} shows Mean IOU of the predicted segmentation on test data, where PSWA provides significant improvement over SGD based training. 

\section{Reproducibility}\label{sec:reproduce}
We provide details of hyperparameter values and additional implementation details about Experiments section.

For the experiments on Cifar10
We adopted the implementation in (\url{ https://github.com/kuangliu/pytorch-cifar/blob/master/main.py}) with the exception of a custom learning rate schedule as the original was sub-optimal. Data augmentation on the training set was performed using random crop (padding 4) and horizontal flip while both train and test were normalized. The dataloaders perform random shuffle on data batches, with 2 concurrent workers for each test and train data queue. The code runs on Pytorch 1.0, Python 3.6, CUDA 9.0 with cuDNN. We use 1 Tesla V100 with 16 GB GPU memory and 8vCPU Intel Skylake.

For experiments on ImageNet
We used the implementation in (\url{ https://github.com/pytorch/examples/tree/master/imagenet}). We used a batch size of 32, L2 penalty of 0.0001, momentum of 0.9, and perform standard data augmentation like cropping, horizontal flipping and input data normalization. The code runs on Pytorch 1.0, Python 3.6, CUDA 9.0 with cuDNN. We use 4 Tesla V100 with 64 GB GPU memory and 32vCPU Intel Skylake.

For experiments on human pose estimation
We adopted the implementation in \\ (\url{ https://github.com/Microsoft/human-pose-estimation.pytorch}) .
The code runs on Pytorch 1.0, Python 3.6, CUDA 9.0 with CudNN. We use 4 Tesla V100 with 64 GB GPU memory and 32vCPU Intel Skylake.

For semantic segmentation experiments
We used \\ (\url{https://github.com/CSAILVision/semantic-segmentation-pytorch}) for implementation details.
The code runs on Pytorch 1.0, Python 3.6, CUDA 9.0 with cuDNN. We use 4 Tesla V100 with 64 GB GPU memory and 32vCPU Intel Skylake.

\section{Loss Surface Analysis}\label{sec:analysis}

\begin{figure*}
\centering
\begin{tabular}{cccc}
Model&
Epoch 20  &
Epoch 100 &
Epoch 150 \\
 & 
 (Early Training) &  (Near Convergence) & (At convergence)  \\
\hline
SGD& 
\includegraphics[width=3.8cm]{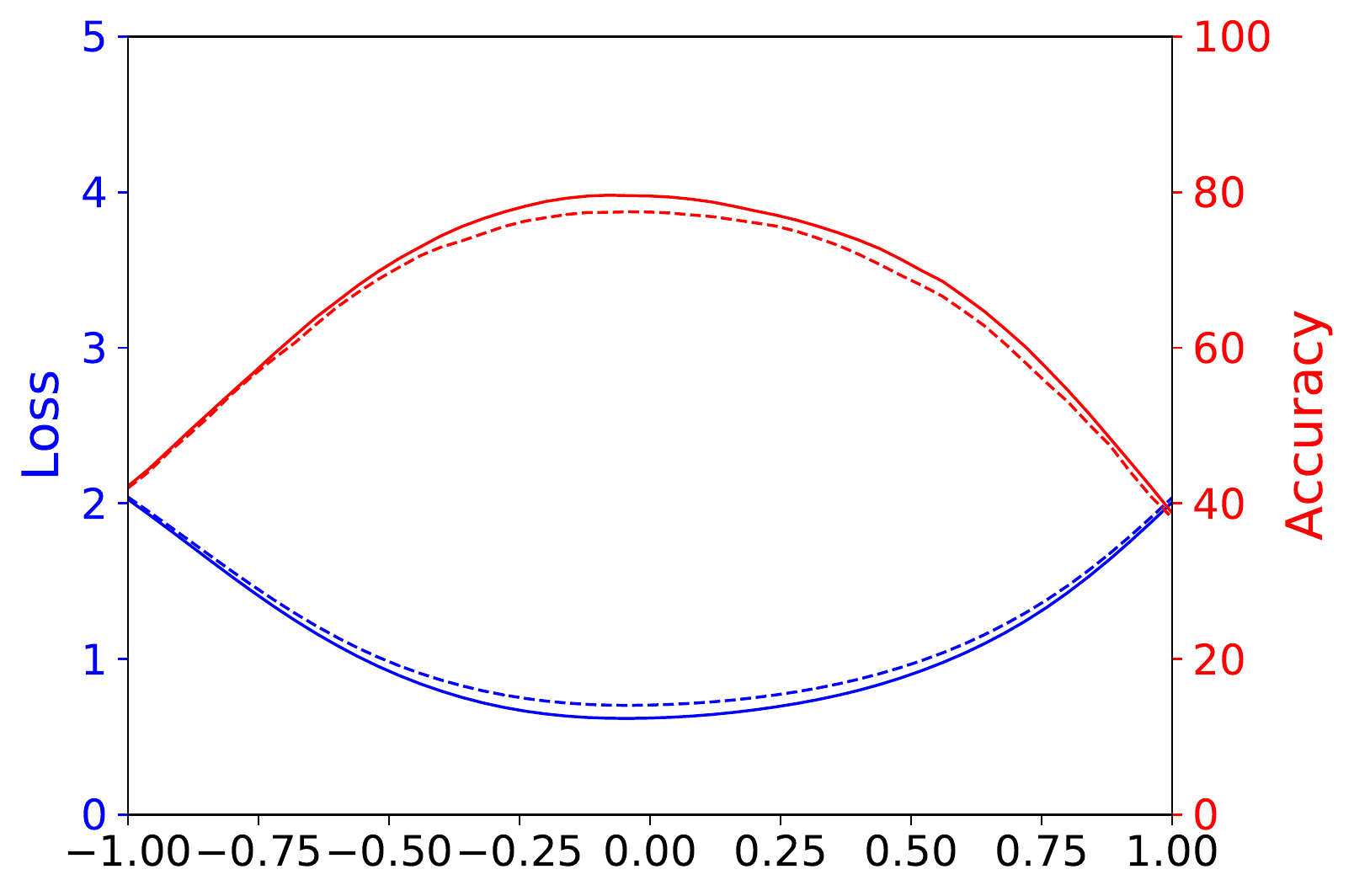} &
\includegraphics[width=3.8cm]{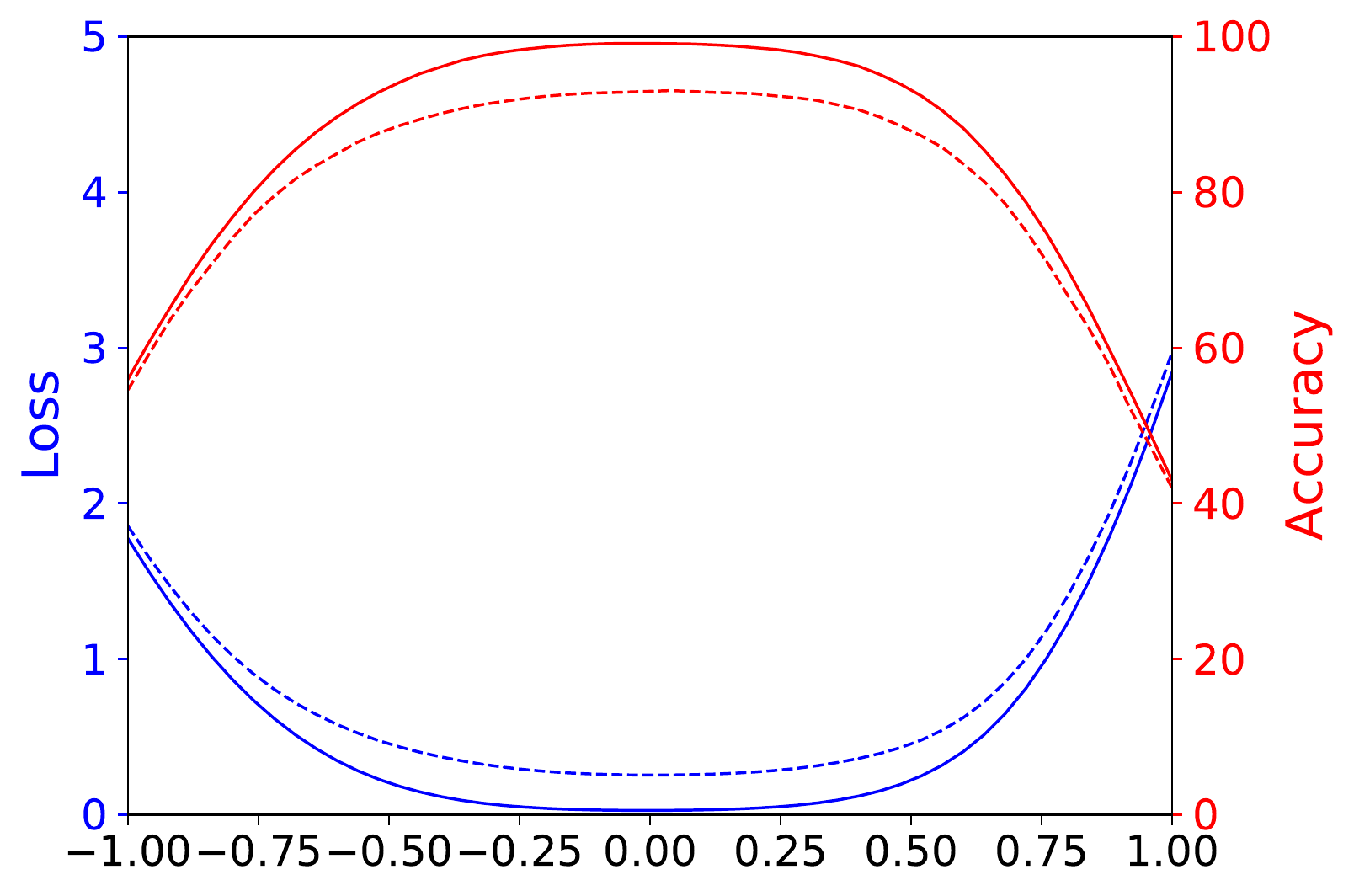} &
\includegraphics[width=3.8cm]{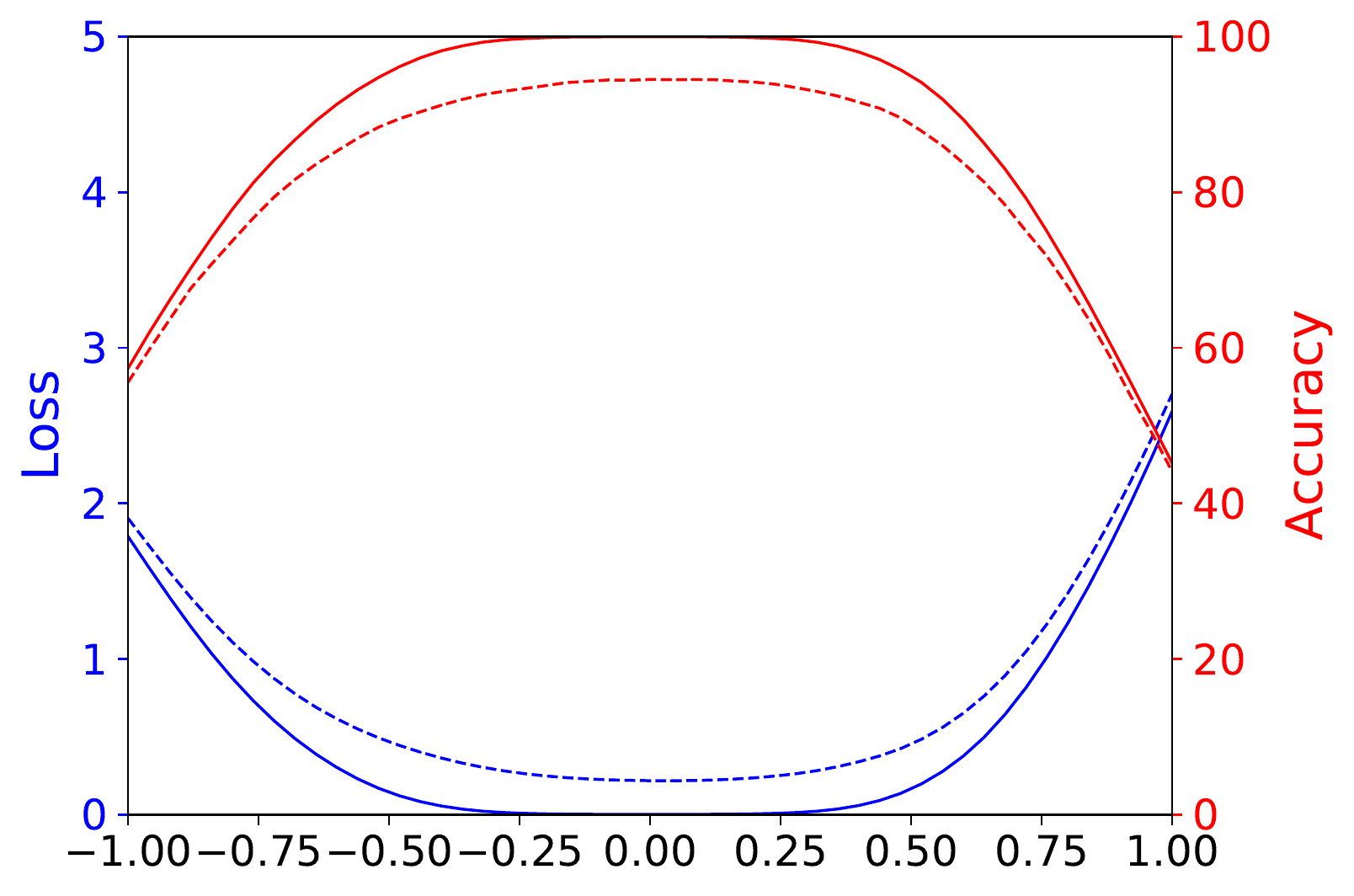} \\
SGD + PSWA &
\includegraphics[width=3.8cm]{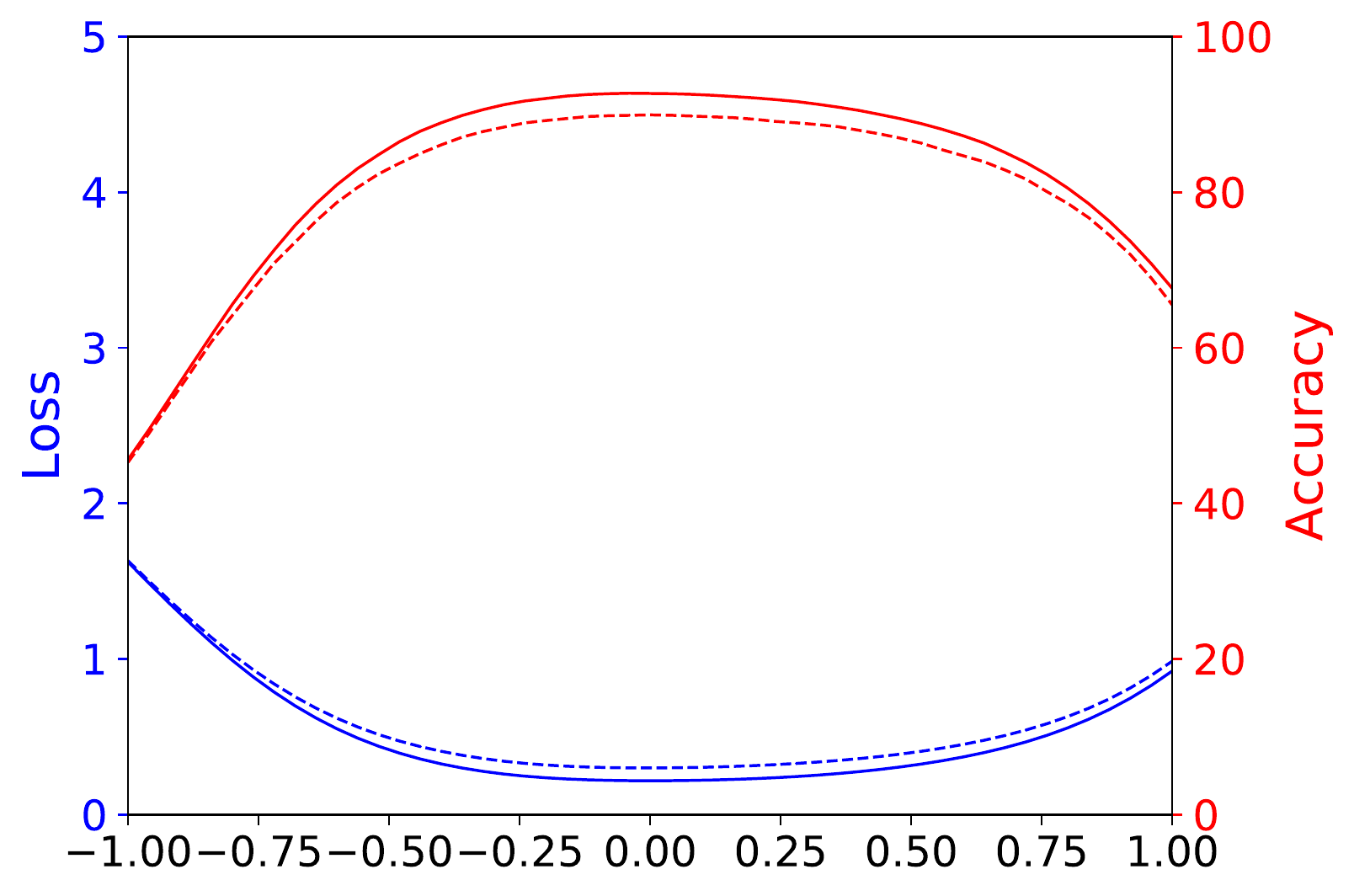} &
\includegraphics[width=3.8cm]{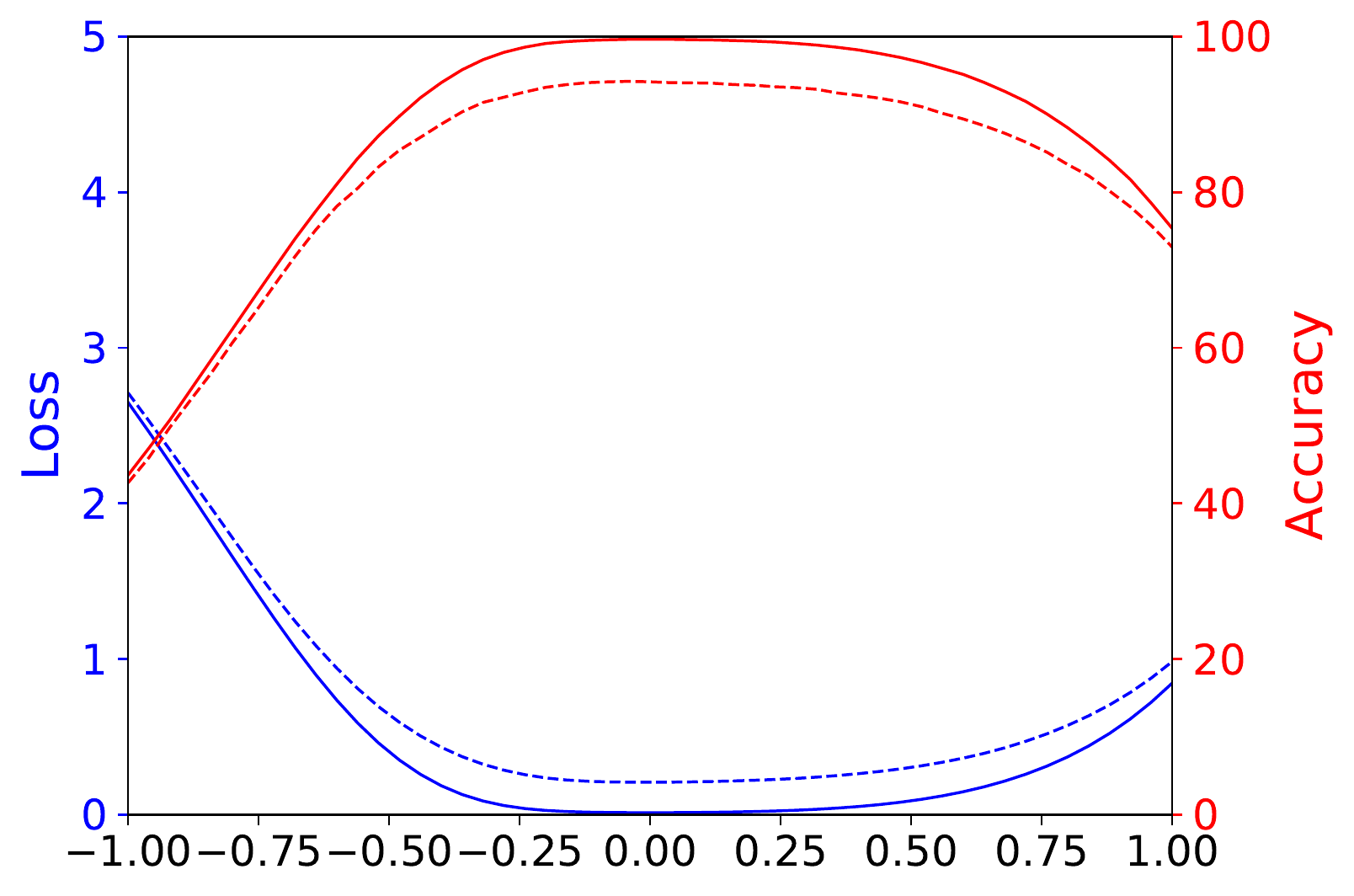} &
\includegraphics[width=3.8cm]{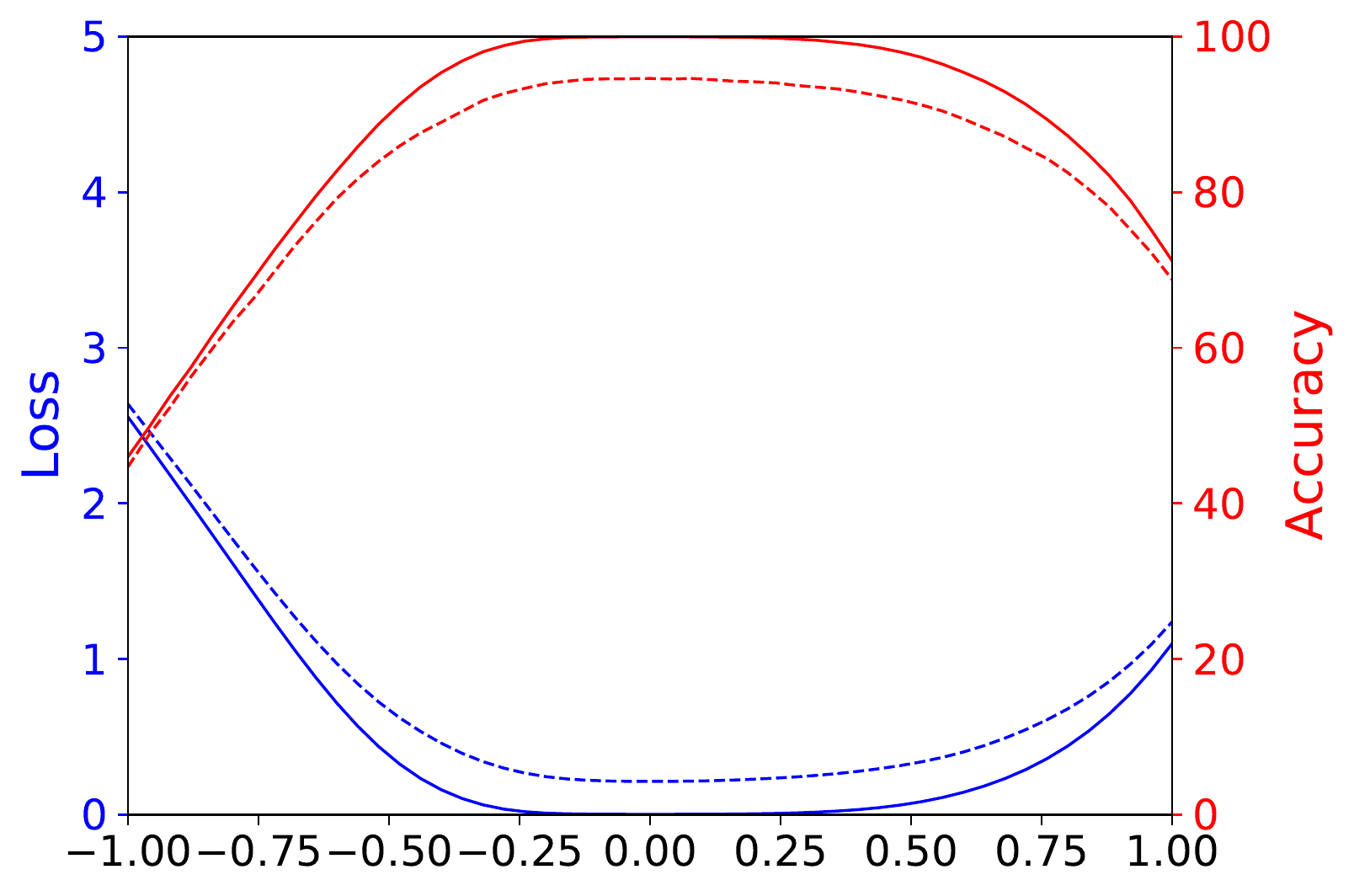}\\
\\
\end{tabular}
\caption{Comparison of loss surface with \ac{SGD} and \ac{PSWA} at different training stages.}
\label{fig:cmp_loss}
\end{figure*}

To better understand the results of our approach, we investigate the effect of \ac{PSWA} on the loss surface of the model during training when compared to \ac{SGD}. Training neural networks requires minimizing a high-dimensional non-convex loss function, with a deeper minima correlating with better performance. An important characteristic of the minima is its `flatness' or the measure of size of the connected region around the minimum where the training loss remains low. There exist strong claims that ``flat'' minima generalize better, while increased sharpness of a minima could indicate low generalization (\cite{hochreiter1997flat,kawaguchi2016deep}). \cite{izmailov2018averaging} shows their technique, \ac{SWA} (based on averaging multiple points along the trajectory of \ac{SGD}) leads to solutions corresponding to wider optima than \ac{SGD}. We draw similar conclusions for \ac{PSWA}.  

\begin{figure}
\centering
\includegraphics[width=\columnwidth]{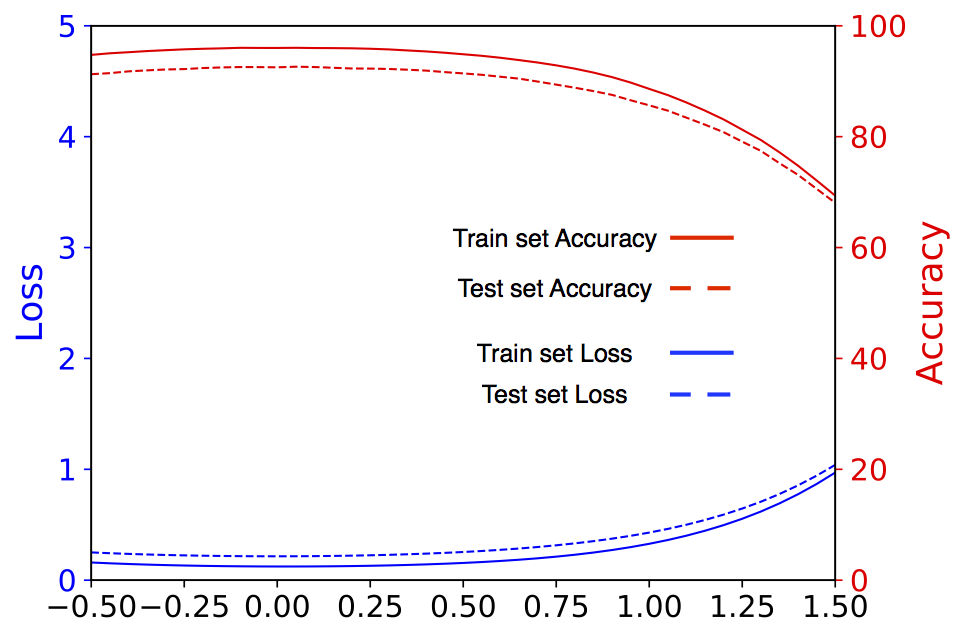}
\caption{Loss surface of model weights before and after applying PSWA on one epoch.}
\label{fig:loss_detail}
\end{figure}

We follow \cite{li2017visualizing} approach, which presents a technique that calculates and visualizes the loss surface along random direction(s) near the weight space. They use a novel ``filter normalization'' scheme that enables side-by-side comparisons of different minima, which addresses problems with 1-Dimensional Linear Interpolation. \autoref{fig:cmp_loss} presents loss surface comparison at different stages of training- beginning, near convergence, and at convergence on ResNet18 for Cifar10 trained by \ac{SGD} and \ac{SGD} with \ac{PSWA}. The horizontal axis represents the displacement of the random Gaussian direction vector; the red lines indicate accuracy and the blue lines indicate the loss values; the dashed lines represent the values on the test dataset while the solid lines represent the training set. As is clearly evident, the model trained by \ac{PSWA} has much flatter and deeper minima, for both training and testing set, at the early training stage. The trend continues for near convergence stage and at convergence, though it becomes less pronounced. We can see that with similar test and train accuracies, PSWA still retains wider minima. \autoref{fig:loss_detail} presents a different representation of the loss surface at early training stage (epoch 50), before and after reassigning the model weights. The \ac{PSWA}-based model is located at index 0 on the horizontal axis, and SGD model at index 1, while variables between them represent the displacement in the ``filter normalized'' direction between the weights (since we use the same model). We notice steady improvements in performance in the direction of weights after PSWA is applied.

{\small
\bibliographystyle{ieee_fullname}

}

\end{document}